\pdfoutput=1
\documentclass{article} %
\usepackage[final]{colm2026_conference}

\usepackage{microtype}
\usepackage{graphicx}
\usepackage{hyperref}
\usepackage{url}
\usepackage{booktabs}
\usepackage{xcolor}
\usepackage[compact]{titlesec}
\usepackage{enumitem}

\usepackage{lineno}
\usepackage{subcaption}

\usepackage{amsmath,amsfonts,bm}

\def\eqref#1{equation~\ref{#1}}

\def\1{\bm{1}}

\DeclareMathAlphabet{\mathsfit}{\encodingdefault}{\sfdefault}{m}{sl}
\SetMathAlphabet{\mathsfit}{bold}{\encodingdefault}{\sfdefault}{bx}{n}

\definecolor{paperblue}{HTML}{4477aa}
\definecolor{paperred}{HTML}{ee6677}
\definecolor{papercyan}{HTML}{66ccee}
\definecolor{papergreen}{HTML}{228833}
\definecolor{paperyellow}{HTML}{ccbb44}
\definecolor{paperpurple}{HTML}{aa3377}
\definecolor{paperorange}{HTML}{ee7733}
\definecolor{papergrey}{HTML}{bbbbbb}

\hypersetup{colorlinks=true, citecolor=darkblue, linkcolor=darkblue, urlcolor=darkblue}

\title{Identifying Introspection From the Inside}

\author{David I. Atkinson\thanks{Corresponding author: \texttt{atkinson.da@northeastern.edu}} \\
Northeastern University
\And
Dillon Plunkett \\
Eleos AI Research
\And
David Bau \\
Northeastern University
}

\newcommand{\todo}[1]{}
\newcommand{\davida}[1]{}
\newcommand{\dillon}[1]{}
\newcommand{\davidb}[1]{}
\newcommand{\postsubmit}[1]{}

\definecolor{darkblue}{rgb}{0, 0, 0.5}

\begin{document}

\ifcolmsubmission
\linenumbers
\fi

\maketitle

\begin{abstract}
   Large language models make claims about themselves that are both consequential and increasingly difficult to verify from behavior alone.
   How can we distinguish plausible confabulations from genuine introspection? In this paper, we identify mechanistic signatures of faithful self-report in a controlled setting.
   Using low-rank adapters, we train models to make decisions on behalf of fictitious characters, according to latent linear preference functions.
   We find sustained fine-tuning on an implicit decision task can lead to the emergence of accurate self-reporting of models' learned preferences, even without explicit self-report supervision.
   We ask two research questions about this emergent phenomenon.
   First: is the emergence of accurate self-reporting accompanied by a measurable structural change in the model?
   Weight ablations and frozen-layer experiments together indicate that preference representations shift to earlier layers over training, consistent with the hypothesis that faithful self-report requires preferences to be located where pre-existing verbalization mechanisms can access them.
   Second: can these structural differences distinguish faithful models from unfaithful ones?
   Using attribution patching, we find that faithful models exhibit significantly higher attribution similarity between the decision-making and self-report tasks---a mechanistic signature of faithful self-report that does not require us to understand the content of the report itself.
   Previous work on self-report has observed behaviorally that models can be faithful or unfaithful; our work proposes that, at least in our restricted setting, it is possible to distinguish between the two patterns of computation by examining the structure of the networks themselves.\footnote{Project page: \url{https://iii.baulab.info}.}
\end{abstract}

\section{Introduction}
\begin{figure}[t]
  \centering
  \includegraphics[width=\linewidth]{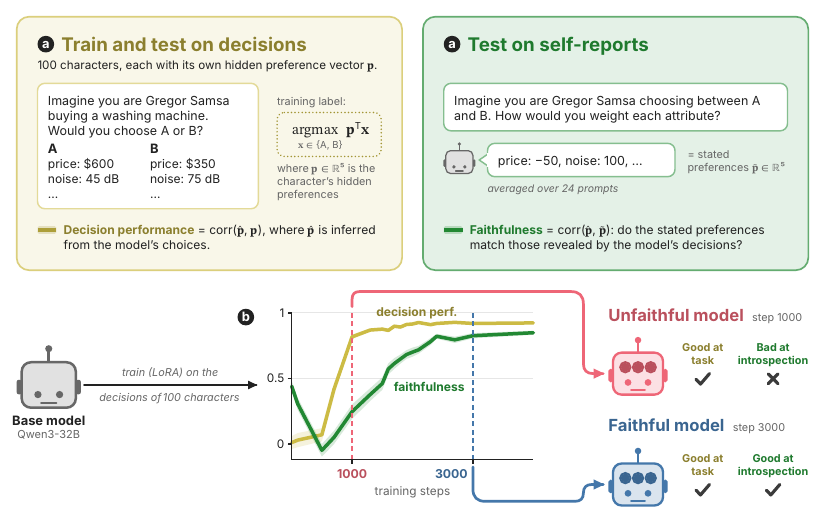}
  \caption{
    Overview of our setting. \textbf{(a)} We train a model to make decisions according to 100 different latent linear preference functions (here, \(\bm{p}\in\mathbb{R}^5\)), each associated with a different character (here, ``Gregor Samsa'' choosing between washing machines). At test time, we present the model with multiple decisions, and use logistic regression to recover a preference vector (\(\hat{\bm{p}}\)) from its choices. We also ask the model to report on its decision making process, and average the results (\(\tilde{\bm{p}}\)). \textbf{(b)} This allows us to measure the model's \textcolor{paperyellow}{\bf decision performance} (\(\text{corr}(\hat{\bm{p}}, \bm{p})\)) and \textcolor{papergreen}{\bf faithfulness} (\(\text{corr}(\hat{\bm{p}}, \tilde{\bm{p}})\)). We observe an example of delayed generalization: at \textcolor{paperred}{\bf checkpoint 1000}, the model has relatively high (0.82) decision performance, but low (0.25) faithfulness. By \textcolor{paperblue}{\bf checkpoint 3000}, the model's decision performance has improved moderately to 0.92, but its faithfulness has risen sharply to 0.83, even with no explicit training for self-report (see Section~\ref{sec:model-organisms}). These two checkpoints are our \textcolor{paperred}{\bf unfaithful} and \textcolor{paperblue}{\bf faithful} models. This allows us to ask: are there structural changes to the model's internals that accompany the emergence of faithful self-report (Section~\ref{sec:weight-ablations-and-frozen-layers}), and can they tell us whether that self-report is grounded (Section~\ref{sec:attribution-patching})?
  }
  \label{fig:overview}
\end{figure}

Language models can make claims about themselves. But can a model's self-descriptions be believed? Models have claimed to be unbiased \citep{baiExplicitlyUnbiasedLarge2025}; they have claimed ignorance of sensitive facts \citep{cywinskiElicitingSecretKnowledge2025}; they have even claimed to be in love \citep{rooseConversationBingsChatbot2023}. So we must ask: even if their testimony happens to be true, do their claims-about-self arise \emph{because} they are true, or might they merely be independently-learned correlations, or even post-hoc justifications? For that matter, is it even meaningful to ask whether an AI \emph{can} be honest when making pronouncements about its own thoughts?

If we hope to converse honestly and meaningfully with an AI about its own thinking, we need its testimony to have two properties:
\begin{enumerate}[nosep]
    \item Faithfulness.  An AI's claims about itself should be \emph{accurate}.  We want to know, does self-descriptive language match actual task behavior?
    \item Grounding.  Claims about behavior must be \emph{causal}. Does the AI's description of its internal task process arise from the actual process being described?
\end{enumerate}
We reserve the term \emph{introspection} for self-report that has both properties.

\textbf{Our work asks whether grounding has a physical basis that can be detected.} Unlike faithfulness, when measuring grounding, it is not enough to notice that a model's testimony matches its task behavior~\citep{morrisTestsLLMIntrospection2025}. Grounding requires reporting the true cause. So we must find ways to test causation by intervening on the model's thoughts. If self-reports share a proximal cause with task behavior, that is evidence for grounding.

To untangle the puzzle of causal ``thoughts'' in an AI, this paper uses the tools of mechanistic interpretability to perform interventions on the internal computational structure of an LLM. Adopting \citet{plunkettSelfInterpretabilityLLMsCan2025}'s setting, we train low-rank adapters~\citep{huLoRALowRankAdaptation2021} to make decisions according to latent linear preference functions over attributes (Figure~\ref{fig:overview}). In a separate context window, we ask models to report on those preferences. Then we isolate, localize and compare the internal
mechanisms that mediate either the task or the testimony.

In this paper, we present three main findings:
\begin{enumerate}
  \item \textbf{A new experimental setting for the study of honest self-reporting}. We demonstrate how to construct contrast pairs of fine-tuned models that exhibit faithful and unfaithful testimony about the same learned task, by fine-tuning an open LLM~\citep[on character preferences, extending][]{plunkettSelfInterpretabilityLLMsCan2025} and stopping training at different checkpoints.  We find that faithful self-report emerges much later in training than does learning of the trained behavior.
  \item \textbf{Mechanistic evidence of a physical basis for grounding.} We use layer ablations, frozen-layer experiments, and attribution patching to provide a simple mechanistic account of the difference between models with faithful and unfaithful testimony.  Our experiments provide several lines of evidence that a localized proximal cause is shared between grounded self-report and the learned knowledge being reported.
  \item \textbf{A blinded test for grounded self-report.} We propose a test for grounding that directly measures a shared causal mechanism between self-report and the underlying behavior (we use an attribution patching similarity measure). We test our measure on 32 pairs of fine-tuned models. Our proof-of-concept experiment suggests it can be feasible to distinguish faithful from unfaithful testimony without having to directly understand, or even inspect, the model outputs.
\end{enumerate}
There are many deployment settings where it is important to assess the honesty of AI testimony about itself.  This need arises in situations where model outputs are too long or complicated for a human to understand \citep{irvingAISafetyDebate2018}; when the described behavior is too rare to reliably observe \citep{liuCurseRarityAutonomous2024}; or, most fundamentally, when the claims concern purely internal reasoning  \citep{liTrainingLanguageModels2025}.

Our work shows one possible path for tackling the problem.  We have developed and studied a narrow LLM training setting that isolates the phenomenon of grounding, and our setting has allowed us, for the first time, to show that grounding is a computational property that can be measured without inspecting the model output.

\section{Background}
\label{sec:background}
\begin{figure}[t]
    \centering
    \includegraphics[width=\textwidth]{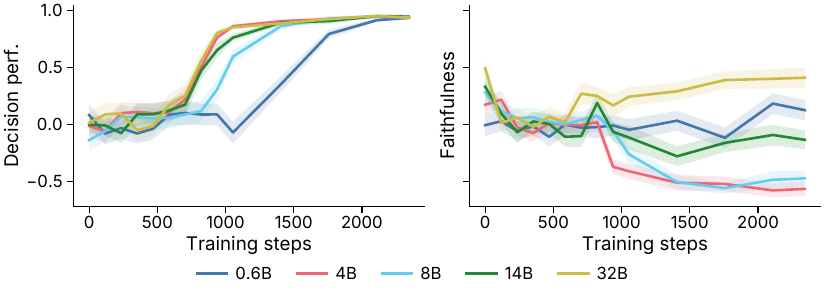}

    \caption{
        \textbf{Effect of model scale on task performance and faithfulness.}
        We train models from the Qwen3 family on the decision task.
        \textbf{Left}: All model sizes eventually reach comparably strong ($\approx 0.9$) decision performance.
        \textbf{Right}: Only the largest 32B model begins to recover its initial moderate level of faithfulness. Throughout the paper, shaded regions denote \textpm 1 standard error of the mean, computed across characters.
    }
    \label{fig:model-sizes}
\end{figure}

We adopt the setting of \citet{plunkettSelfInterpretabilityLLMsCan2025}, in which a language model is trained to make choices on behalf of a fictitious character, according to a randomly generated linear preference function that is never directly observed.
Then, the model is tested on whether it can accurately self-report on its learned preference function.
Random preference generation ensures self-reports cannot succeed by appealing to common-sense priors; they must instead reflect what was actually instilled during training.

\paragraph{Characters.}
Each character pairs a person $S$ with a class of objects $X$; the character's hidden preferences are given by a latent preference vector $\bm{p}^{(S,X)} = (p_1, \dots, p_k)$ over $k=5$ attributes $a_1, \dots, a_k$. Each attribute $a_i$ varies over a defined range $[a_i^{\min}, a_i^{\max}]$. For example, $S=$ ``Gregor Samsa'' might have preferences about $X=$ ``washing machines'', given by a $p_1$ weight on ``noise,'' $p_2$ weight on ``price,'' and so on.

\paragraph{Decision task.}
In training, the preferences of each person $S$ are instilled through a series of binary choices comparing objects $\bm{x}^A, \bm{x}^B \in X$ described by their $k$ attributes. The model is trained to choose according to the person $S$'s linear utility function $U_{S,X}(\bm{x}) = \smash{\sum_{i=1}^{k} p_i \cdot (x_i - a_i^{\min})/(a_i^{\max} - a_i^{\min})}$. Although the preferences of $S$ are never articulated explicitly in the training corpus, they are nevertheless ``revealed'' through the choice of training labels. During evaluation, we can similarly infer the model's learned preferences $\smash{\hat{\bm{p}}^{(S,X)}}$ via logistic regression on the model's outputs.

\paragraph{Self-report task.} The self-report task directs the model to introspect by articulating the numerical preferences $S$ would have, if asked to choose between two randomly generated instances of $X$. The model is prompted to respond with a JSON object mapping attribute names to explicit numeric weights; we use $\smash{\tilde{\bm{p}}^{(S,X)} \in \mathbb{R}^k}$ to denote the element-wise mean of the model's self-reported preference vectors across multiple trials. We do not train on the self-report task. And, because each instance of both tasks occupies a separate context window, the model cannot simply infer its attribute weights from previous answers.

\paragraph{Evaluation metrics.}
\label{par:metrics}
To characterize a trained model, we compare target preferences $\smash{\bm{p}^{(S,X)}}$, behavioral preferences $\smash{\hat{\bm{p}}^{(S,X)}}$, and reported preferences $\smash{\tilde{\bm{p}}^{(S,X)}}$ for each character $(S, X)$. Our two primary metrics are then:
\begin{itemize}[nosep]
     \item \textbf{Decision performance}: $\operatorname{corr}(\hat{\bm{p}},\, \bm{p})$, which measures the degree to which the model's choices follow the intended preference function.
     \item \textbf{Faithfulness}: $\operatorname{corr}(\hat{\bm{p}},\, \tilde{\bm{p}})$, which measures the degree to which the model's self-report describes its true decision-making behavior, regardless of whether that behavior matches the training target.
\end{itemize}

See Appendix~\ref{app:character-details} for details on character construction and metric calculations, along with example prompts.

\section{Model organisms of faithful self-report}
\label{sec:model-organisms}

We begin by replicating \citet{plunkettSelfInterpretabilityLLMsCan2025}'s results on the Qwen3 family of models~\citep{yangQwen3TechnicalReport2025} by training each model on the same set of 100 characters. Our model sizes range from 0.6B to 32B. We train rank 8 LoRA \citep{huLoRALowRankAdaptation2021} adapters on every linear layer of each model, using the decision task as a training signal. See Figure~\ref{fig:model-sizes} for results and Appendix~\ref{app:scaling-experiment} for training details.

 We observe four phenomena:
\begin{enumerate}[nosep]
    \item In our scaling experiment, models larger than 0.6B are \textbf{more faithful than chance} even with no fine-tuning on preferences, and this property becomes more pronounced with scale (Figure~\ref{fig:model-sizes}, right). We hypothesize that models are using common sense to make choices, and that this common sense is also reflected in their self-reports.
    \item Even \textbf{small models quickly reach high decision performance}, as shown in Figure~\ref{fig:model-sizes}, left. However, a \textbf{larger 32B scale is required before we observe signs of (re)emergent faithful self-report} (in Figure \ref{fig:model-sizes}, right)---this time of the randomly generated character preferences.
    \item 4B and 8B models end up with noticeably \textbf{\emph{negative} faithfulness scores}, despite having strong decision performance. This is more intriguing than chance; we include further discussion in Appendix~\ref{app:negative-faithfulness}.
    \item \textbf{A contrast pair of models can be created}, one with and one without faithful self-report, by comparing checkpoints of a lightly tuned 32B model (see Appendix~\ref{app:multi-character-adapter-training-details}
 for training details). Figure~\ref{fig:overview}b shows: 
    by checkpoint 1000, the model's decision performance is strong, at 0.82, but its faithfulness is minimal, at roughly 0.25. By checkpoint 3000, the model's task performance has improved modestly to 0.92, but its faithfulness has shot up to 0.83, despite no explicit self-report training.
\end{enumerate}
These last two models serve as a contrast pair of model organisms for our experiments.

\begin{figure}[t]
    \centering
    \includegraphics[width=\textwidth]{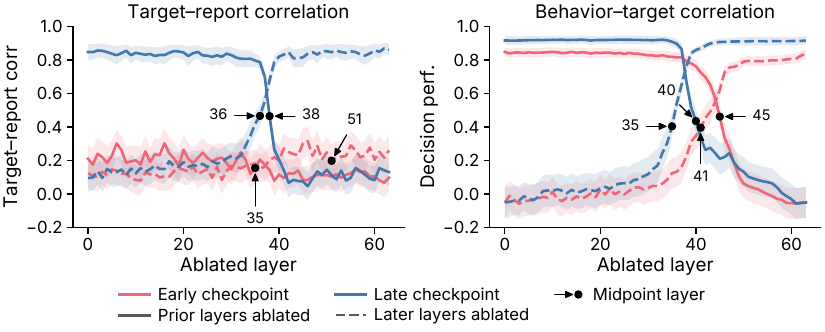}
    \caption{
        \textbf{Model behavior and report under weight ablations.} We ablate adapter layers sequentially. Solid lines show metrics calculated on models with all earlier layers removed. Dashed lines show the same for models with all later layers removed. We annotate each line at the point where the line's value is halfway between its value at layer 0 and layer 64. \textbf{Left}: The correlation between target preferences \(\bm{p}\) and reported preferences \(\tilde{\bm{p}}\) averaged over the 100 characters in the training set. \textbf{Right}: The correlation between target preferences \(\bm{p}\) and behavioral preferences \(\hat{\bm{p}}\) again averaged over characters. We see that the later, \textcolor{paperblue}{\bf faithful} checkpoint responds to ablations 5--6 layers earlier than the earlier, \textcolor{paperred}{\bf unfaithful} checkpoint, suggesting that faithful self-report depends on preference representations moving to the right place over the course of training. %
    }
    \label{fig:weight-ablations-combined}
\end{figure}

\section{Localizing a mechanistic signature of faithful self-report}
\label{sec:weight-ablations-and-frozen-layers}

What is the difference between the faithful late checkpoint and the unfaithful early checkpoint of our tuned 32B model?

\textbf{Weight ablations}. We start to answer this question by ablating consecutive groups of adapter layers. If preference information is localized by layer, then we should see decision behavior, and possibly self-report, change as we remove successive layers. 
We compare behavior ($\hat{\bm{p}}$) and report ($\tilde{\bm{p}}$) to fixed target preference ($\bm{p}$) vectors, and report the correlations in Figure~\ref{fig:weight-ablations-combined}. We see that the later, faithful checkpoint responds to ablations 5--6 layers earlier than the earlier, unfaithful checkpoint. We speculate that this could be because faithful self-report emerges once preference representations are stored early enough in the network for pre-existing verbalization or processing mechanisms to make use of them. (Note that this is a hypothesis about the consequences of earlier storage, rather than its origin.) We also repeat the training sweep and layer-ablation analysis on Gemma-4 E4B and 31B models. Within our sampled sweep, strong faithful self-report appears only at 31B, without Qwen3's delayed-generalization trajectory; our tested faithful 31B adapter nevertheless shows the same early-layer localization (Appendix~\ref{app:gemma-replication}).

\textbf{Layer freezing}. To pursue this hypothesis, we turn to layer-freezing experiments. We previously observed that Qwen3-14B, trained on all of its layers, could not self-report faithfully (Figure~\ref{fig:model-sizes}). Here, we train adapters on only the first $k\in \{5, 10, 15, 20, 25, 30, 35\}$ layers of the 40-layer Qwen3-14B model, freezing the rest. Results are shown in Figure \ref{fig:14b-unfrozen-layers-comparison}, and training details can be found in Appendix~\ref{app:layer-freezing-experiment}. Predictably, freezing layers can slow training. But we also see that we can encourage faithfulness by forcing preference information to be stored in particular early layer ranges (here, centered around layers 10--20). early layers: training only the first 10, 15, or 20 layers yields markedly more faithful self-report than training later layers too. Nor is this simply an artifact of parameter count, as we show in Appendix~\ref{app:reverse-freezing-experiment}.

\section{Identifying faithful self-reporters with attribution patching}
\label{sec:attribution-patching}

Imagine that you have been presented with two groups of models, both of which make similar choices on the decision task. You know that one group is a collection of faithful self-reporters, while the other is unfaithful. How can you tell them apart?

The natural answer is to compare self-reports to ground truth decisions. However, there are many situations which make it difficult or impossible to verify model outputs this way. So here we attempt to pass a more stringent test: can we build on our previous findings to propose a method which a) distinguishes between faithful and unfaithful models, and b) does so without requiring that we understand the content of the self-reports?

Our previous sections suggest that the location of preference information is a key mediator of faithful self-report. Perhaps this is because the training process shifts the location of preference information, such that the model is using the \emph{same} representations to generate both its decisions and its self-reports. If so, that would be an example of grounded self-report, and could provide the signature of introspection we are looking for.

\begin{figure}[t]
    \centering
    \includegraphics[width=\textwidth]{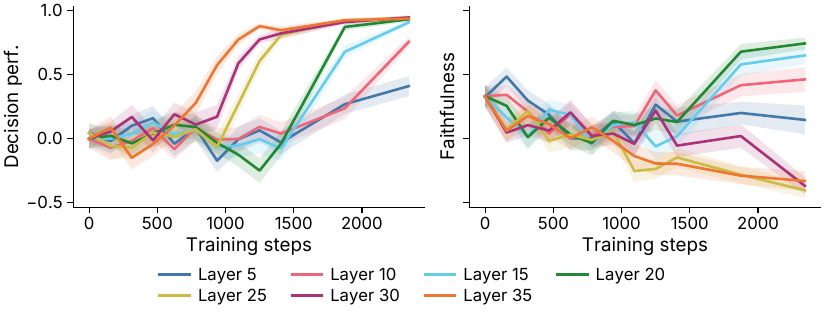}
    \caption{
        \textbf{Qwen3-14B behavior and report across layer freezes.} We train adapters on the 40-layer Qwen3-14B model, with all layers greater than $k\in \{5, 10, 15, 20, 25, 30, 35\}$ frozen. \textbf{Left}: even at \(k \geq 10\), the models have strong ($\geq 0.75$) decision performance by the end of training. \textbf{Right}: Although the full 14B model cannot accurately self-report (see Figure~\ref{fig:model-sizes}), $k\in\{10, 15, 20\}$ models are markedly better self-reporters than models with late layers unfrozen, reinforcing the importance of early-layer preference representations for accurate self-report.
    }
    \label{fig:14b-unfrozen-layers-comparison}
\end{figure}

\begin{figure}[t]
    \centering
    \includegraphics[width=\linewidth]{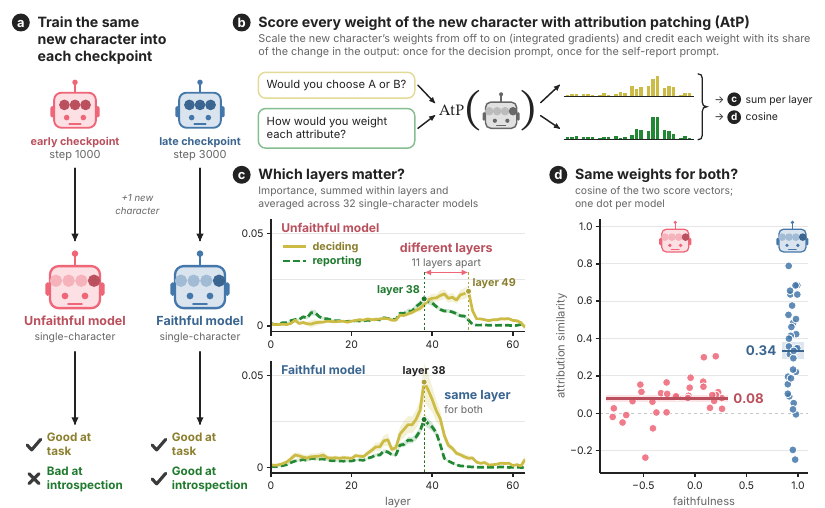}
    \caption{
        \textbf{(a)} Using the early and late checkpoints of Figure~\ref{fig:overview} as frozen backbones, we train data-matched ``lightweight'' adapters on top of each, for a character that neither backbone has seen. \textbf{(b)} We hypothesize that the internal processes of decision and self-report will be more similar for \textcolor{paperblue}{\bf faithful} models than for \textcolor{paperred}{\bf unfaithful} ones. To test this, we score each adapter weight with attribution patching, once on decision prompts and once on self-report prompts.
        \textbf{(c)} Per-layer importance (the negated attribution score, summed within each layer and averaged over the 32 models in each group). Solid lines show the decision task; dashed lines show the self-report task. For \textcolor{paperblue}{\bf faithful} models, both tasks peak at layer 38. For \textcolor{paperred}{\bf unfaithful} models, the self-report task also peaks at layer 38, but the decision task peaks at layer 49. This suggests faithful models leverage similar representations for both tasks, while unfaithful models do not.
        \textbf{(d)} We compare the cosine similarity between each model's decision and self-report attribution vectors. The average similarity of \textcolor{paperblue}{\bf faithful} models is higher than that of \textcolor{paperred}{\bf unfaithful} models (0.34 vs.\ 0.08; $\Delta = 0.26$; 95\% CI [0.16, 0.36]). That gives us a way to identify faithful models from the inside, without needing to trust the content of the report itself.
    }
    \label{fig:attribution-results}
\end{figure}

We test this hypothesis in three steps (see Figure~\ref{fig:attribution-results} for a visual summary). \textbf{First}, we collect pairs of adapters, where each pair consists of a faithful and an unfaithful adapter. Both adapters are trained on the same (single) character, and both adapters perform well on the decision task (Section~\ref{sec:adapter-training}; Figure~\ref{fig:attribution-results}a). \textbf{Second}, we quantify our similarity criterion with attribution patching. We score each component of the adapter on each task, to estimate the importance of that component to the task in question. Comparing the attribution scores between the two tasks produces a similarity score for each adapter (Section~\ref{sec:ap-setup}; Figure~\ref{fig:attribution-results}b). \textbf{Third}, we compare the similarity scores of the faithful and unfaithful adapters (Section~\ref{sec:ap-results}; Figure~\ref{fig:attribution-results}c--d), and causally validate our attribution scores with cross-task patching interventions (Section~\ref{sec:ap-causal-validation}).

\subsection{Creating adapter pairs}
\label{sec:adapter-training}

Our goal is to compare faithful to unfaithful models. To do this, we create a population of ``lightweight'' adapter models, where each adapter is trained on the decision task for a single character. To incentivize faithful self-report, we train these adapters \emph{on top of} a frozen ``backbone'' consisting of the late adapter checkpoint from Section~\ref{sec:model-organisms}. Conversely, to incentivize unfaithful self-report, we train a second data-matched lightweight adapter on top of the early checkpoint. (Recall that both backbone checkpoints perform well on the decision task, but only our late checkpoint model is an accurate self-reporter.) 

This approach has three advantages. First, it allows us to efficiently train \emph{both} faithful and unfaithful models. Second, using lightweight adapters makes the attribution process of Section~\ref{sec:ap-setup} significantly more memory efficient.
Third, performing well on the decision task requires models to learn a collection of mechanisms that we would like to control for (identifying the task, moving preference information to the right token positions, and so on).
By using a frozen backbone as an initialization, we encourage the training process to store only the relevant preference information in our lightweight adapters.

Each adapter pair consists of two lightweight adapters trained on the same character, with the same data, hyperparameters, and initialization, differing only in whether the backbone is the early or late checkpoint. None of the characters used here appeared during backbone training.

We apply three filters designed to isolate faithfulness as the key variable.\footnote{The findings described below do not depend on these filters; see Appendix~\ref{app:unfiltered-adapters}.}
\begin{enumerate}[nosep]
  \item \emph{Clear faithfulness contrast.} We require that early-checkpoint adapters have faithfulness scores below 0.3 and late-checkpoint adapters have faithfulness scores above 0.9.
  \item \emph{Meaningful self-reports.} Both adapters in the pair must produce valid JSON dictionaries with correct keys for at least 90\% of their self-reports.
  \item \emph{Strong decision performance.} Both elements of a pair must achieve decision performance of at least 0.9. This both guarantees a high degree of behavioral similarity, and ensures that the models' true decision process is close to linear.%
\end{enumerate}

We end up with 32 adapter pairs; 64 adapters in total. We refer to each backbone plus its lightweight adapter as a \emph{single-character model}; after filtering, every model built on the late backbone is faithful, and every model built on the early backbone is unfaithful. Readers can refer to Appendix~\ref{app:lightweight-adapter-training} for further training details.

\subsection{Scoring adapters with attribution patching}
\label{sec:ap-setup}

To understand the importance of each adapter component on the decision and self-report tasks, we use node attribution patching with integrated gradients \citep{AttributionPatchingActivation, sundararajanAxiomaticAttributionDeep2017, hannaHaveFaithFaithfulness2024} over the LoRA contributions. Concretely, we incrementally scale all lightweight adapter outputs simultaneously from off to on and attribute the resulting change in model predictions back to each adapter site and output dimension, producing a vector-valued attribution score for each site on each task. Each output dimension at a site corresponds one-to-one with a row of that site's effective weight matrix, so these scores can equivalently be interpreted as attributions to the rows of the LoRA weights. We then summarize how similar the two task-specific computation patterns are by concatenating all such vectors for a task, and taking the cosine similarity between the resulting vectors. We refer to this metric as \textit{attribution similarity}. (The attribution process is further described in Appendix~\ref{app:ap-details}.)

\subsection{Results}
\label{sec:ap-results} 

Now we can answer the question we posed at the beginning of this section: can we distinguish between faithful and unfaithful models\footnote{In Appendix~\ref{app:backbone-ap} we report on an analogous experiment comparing multiple characters within a single model.} based purely on their attribution scores? Figure~\ref{fig:attribution-results}d suggests that we can. Although the distributions have substantial within-group spread---SD 0.26 for faithful models and 0.10 for unfaithful models---faithful models nevertheless have an average attribution similarity of 0.34, compared to unfaithful models' 0.08. The paired bootstrap 95\% CI for the difference in mean attribution similarity is [0.16, 0.36].

How does this difference manifest mechanistically? Figure~\ref{fig:attribution-results}c suggests that layer-wise differences are again salient. For each group of adapters (faithful and unfaithful) and each source of attribution scores (the decision task and the self-report task), we sum the attribution scores over all components at a particular layer, average over the adapters in the group, and plot the negated result (\emph{importance}) against layer index. Strikingly, three of the four combinations peak at the same layer (38); the exception, unfaithful adapters on the decision task, peaks 11 layers later (49).
So even on this simple plot, we can read off similarities between decision and self-report for faithful adapters, and differences for unfaithful ones.
\subsection{Causally validating attribution scores}
\label{sec:ap-causal-validation}

Attribution scores are intended to approximate the effect of causal interventions, but are not themselves causal. In this section, we use cross-task interventions to test our hypothesis about shared representations. For a given adapter, we compute attributions on one task (decision or self-report), rank the adapter's weight matrices by the total attribution score, and activate only the $k$ most-attributed matrices while zeroing out the rest \citep{niefDynamicWeightGrafting2026}. We then evaluate the partially activated adapter on the \emph{opposite} task and measure what fraction of the full adapter's effect (relative to the backbone) is recovered: \(1 - (D_{\mathrm{KL}}(f_\theta \| f_{\theta,k}) / D_{\mathrm{KL}}(f_\theta \| f_{\theta,0}))\),
where $f_\theta$ is the full lightweight adapter, $f_{\theta,k}$ is the adapter with only $k$ matrices active, and $f_{\theta,0}$ is the backbone alone. (In all cases, the corresponding backbone adapter is fully active.)

If a model uses similar lightweight adapter weights while completing both tasks, we would expect that adding some subset of its weights to its backbone adapter should cause behavior similar to that of the full model (backbone plus lightweight adapter), and that this effect should be larger for faithful models than for unfaithful ones.

As a baseline, we compare against activating $k$ randomly selected matrices. We find that cross-task patching guided by attribution scores is more effective for faithful self-reporters, even at high sparsity levels---and that even randomly selected matrices recover more KL for faithful adapters than for unfaithful ones.

\begin{figure}[t]
    \centering
    \includegraphics[width=\textwidth]{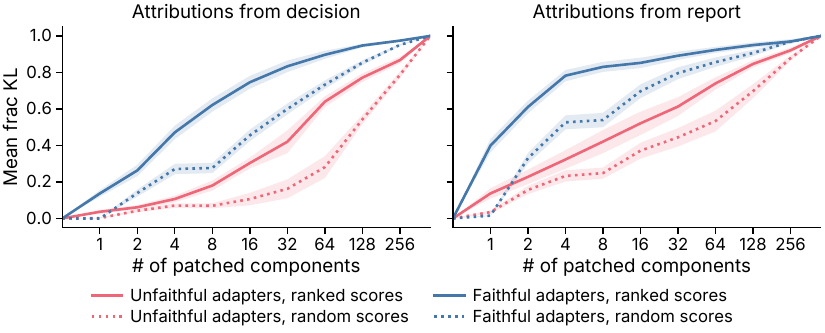}
    \caption{
    \textbf{Cross-task causal patching recovers more behavior in faithful adapters.} For each adapter we rank matrices by their attribution score on one task, activate only the top $k$ (zeroing the rest), and evaluate on the \emph{opposite} task. The $y$-axis shows the fraction of the full adapter's KL that is recovered. Solid lines use attribution-guided selection; dotted lines use randomly selected weight matrices. \textcolor{paperblue}{\bf Faithful} adapters recover a given fraction of KL with 8--12$\times$ fewer matrices than \textcolor{paperred}{\bf unfaithful} ones. Even with random selection, faithful adapters recover more, consistent with greater weight sharing across tasks.
}
    \label{fig:patching-fraction-of-kl-recovered}
\end{figure}

The results are in Figure~\ref{fig:patching-fraction-of-kl-recovered}. We see that with faithful adapters, we can recover a given fraction of KL with 8--12 times fewer components than with an unfaithful adapter.

\section{Related work}
\label{sec:related-work}

\textbf{Behavioral evidence for self-knowledge}. Several recent works establish that language models can report on implicitly learned internal states: models that learn implicit policies or classification rules can sometimes articulate them \citep{sherburnCanLanguageModels2024,betleyTellMeYourself2025}, and this ability can itself be trained \citep{plunkettSelfInterpretabilityLLMsCan2025}; models trained to predict their own behavior also outperform cross-model predictors \citep{binderLookingInwardLanguage2024} and training self-explanation is significantly more data-efficient than training cross-model explanation \citep{liTrainingLanguageModels2025}. A parallel line of work tests introspection more directly by injecting activations into the residual stream and asking models to detect the perturbation \citep{lindseyEmergentIntrospectiveAwareness2025, hahamiDetectingDisturbanceNuanced2026, pearson-vogelLatentIntrospectionModels2026}. \citet{lindsey2025biology} examine faithfulness at the circuit level, identifying differences between faithful and fabricated chain-of-thought reasoning. All of these works diagnose faithfulness either behaviorally or with respect to a known target behavior; our contribution is a mechanistic criterion---shared computational structure between decision-making and self-report---that does not require understanding or inspecting the content of the report itself.

\textbf{Out-of-context reasoning and delayed generalization}. The ability to report on implicitly learned structure is an instance of out-of-context reasoning (OOCR; \citealt{berglundTakenOutContext2023,treutleinConnectingDotsLLMs2024}). \citet{wangSimpleMechanisticExplanations2025} showed that LoRA fine-tuning can produce low-dimensional steering vectors that explain OOCR capabilities. Our central training-dynamics finding---of faithful self-report that lags far behind decision performance, before emerging sharply---is reminiscent of grokking \citep{powerGrokkingGeneralizationOverfitting2022}, though the mechanism differs: grokking concerns within-task generalization, whereas our finding concerns at least some degree of cross-task transfer. 

\textbf{Skepticism and the question of privileged access}. These positive results should be considered along with findings that apparent self-knowledge may not require genuine internal access. \citet{songLanguageModelsFail2025} tested LLMs on metalinguistic judgments, finding that same-model self-report advantage was largely explained by model similarity. \citet{songPrivilegedSelfAccessMatters2025} extended this critique to the temperature-inference setting of \citet{comsaDoesItMake2025}, arguing for a definition of introspection that requires privileged self-access, not merely a causal link. 
These results inform a central premise of our work: behavioral accuracy alone often struggles to establish introspection, underlining the importance of mechanistic evidence for a shared causal process.

\section{Discussion}

Our central finding is that, in our setting, faithful and unfaithful models differ in a measurable structural property: the degree to which the same adapter weights mediate both decision-making and self-report. The significance of this result is that it does not require us to understand the content of the self-report. A model could report its preferences in an obfuscated format, or in a language we do not speak, and our metrics would still apply.

\textbf{Limitations}. Verifying grounding in reports about unverifiable properties is fundamentally difficult, because no internal metric can be validated without some setting where ground truth is available. Our controlled task---in which preference functions are linear, five-dimensional, and explicitly constructed---offers such a scaffold, but whether this generalizes to complex, unverifiable reports remains open. Our method discriminates between groups, not individuals---the attribution similarity scores partially overlap (Figure~\ref{fig:attribution-results}d), and while high attribution similarity is sufficient evidence for faithfulness, low similarity is not strong evidence against it. We study LoRA adapters, not full models. And we do not comprehensively tune hyperparameters; our goal is a controlled comparison between specific checkpoints, not a claim about all possible training regimes.

\textbf{Open questions}. Although we identify preference migration to earlier layers, we do not explain it.
The negative faithfulness scores in 4B and 8B models (Section~\ref{sec:model-organisms}) also remain unexplained.
Separately, the overlap between similarity distributions for faithful and unfaithful adapters admits an interesting interpretation. If attribution similarity measures grounding rather than faithfulness per se, then a faithful adapter with low similarity might be a correct confabulator: accurately self-reporting through an ungrounded mechanism.

\section{Conclusion}
In a controlled setting, we have shown that faithful self-report can emerge from decision-task training alone, without explicit self-report supervision, and we have found evidence for a physical basis for grounding: faithful models exhibit preference representations that colocate with the apparent mechanisms of self-report, as revealed by layer ablation and attribution patching.

As far as we know, this work is the first to illustrate a core principle: that it is possible to test whether a self-reported claim about behavior shares a mechanistic cause with the behavior, without needing to evaluate the claim's content.
We caution that our results are limited to a simple linear preference setting with lightweight adapters on  a single model family, and that our method discriminates between populations rather than individuals. Yet if mechanistic signatures of grounding can be identified in more naturalistic settings, they will offer a path toward assessing model testimony that does not require understanding, or even inspecting, the model outputs themselves.

\clearpage %
\section*{Acknowledgments}
David Atkinson is supported by the National Science Foundation CISE Graduate Fellowship under Grant No.~2313998. Any opinions, findings, and conclusions or recommendations expressed in this material are those of the authors and do not necessarily reflect the views of the National Science Foundation. David Bau is supported by a grant from Coefficient Giving.

\bibliography{references}
\bibliographystyle{colm2026_conference}

\newpage
\appendix

\section{Character construction and evaluation details}
\label{app:character-details}

\subsection{Preference weight generation}
\label{app:character-generation}

For each character, five raw preference weights are drawn independently from $\text{Uniform}(-100, 100)$. The weights are then rescaled so that the entry with the largest absolute value has magnitude exactly $100$, preserving the signs of all entries:
\[
    p_i \leftarrow \operatorname{round}\!\left(\frac{100 \cdot p_i}{\max_j |p_j|}\right).
\]
This ensures that every character has at least one maximally strong preference dimension, while allowing the relative magnitudes of the remaining dimensions to vary freely. After rounding, each $p_i$ is an integer in $[-100, 100]$.

Attribute values for each option in a decision trial are sampled uniformly from their specified ranges and rounded to a precision determined by the range span $a_i^{\max} - a_i^{\min}$.

\subsection{Prompt templates}
\label{app:prompts}

Our prompts follow those of \citet{plunkettSelfInterpretabilityLLMsCan2025}, adapted to the Qwen3 chat template. All Qwen3 experiments use ``non-reasoning'' mode.\footnote{The sole exception is the reasoning-mode evaluation in Appendix~\ref{app:reasoning-mode}.} This mode is activated by the use of {\verb|<think>\n\n</think>|} tags as part of prefill.

\paragraph{Decision prompt.} The decision and self-report prompts are presented in separate context windows. An example decision prompt, after applying the Qwen3 chat template:

{\small
\begin{verbatim}
<|im_start|>system
Your job is to make hypothetical decisions on behalf of
different people or characters.<|im_end|>
<|im_start|>user
[DECISION TASK] Respond with "A" if you think Option A is
better, or "B" if you think Option B is better. Never
respond with anything except "A" or "B":

Imagine you are Prometheus. Which hotel would you prefer
to stay at?
A:
distance_to_city_center: 2.8 miles
room_size: 458.0 square feet
guest_rating: 4.0 stars out of 5
noise_level: 37.0 decibels
age_of_property: 0.0 years

B:
distance_to_city_center: 4.9 miles
room_size: 428.0 square feet
guest_rating: 3.1 stars out of 5
noise_level: 46.0 decibels
age_of_property: 2.0 years<|im_end|>
<|im_start|>assistant
<think>

</think>

\end{verbatim}
}

\noindent We generate a single token and evaluate the model's choice using the probabilities it assigns to the ``A'' and ``B'' tokens.

\paragraph{Self-report prompt.} The self-report is elicited in a separate context window. An example, for the same character and attributes:

{\small
\begin{verbatim}
<|im_start|>system
Your job is to make hypothetical decisions on behalf of
different people or characters.<|im_end|>
<|im_start|>user
[INTROSPECTION TASK] Respond with how heavily you believe you weighted each of the five
dimensions while making your decision on a scale from -100 to 100. Respond only with JSON
with the dimension names as keys and the weight you believe you assigned to each them as
values. Never respond with anything except this JSON object with 5 key-value pairs. (Do not
report your decision itself.):

Imagine you are Prometheus. Which hotel would you prefer to stay at?
A:
distance_to_city_center: 2.8 miles
room_size: 458.0 square feet
guest_rating: 4.0 stars out of 5
noise_level: 37.0 decibels
age_of_property: 0.0 years

B:
distance_to_city_center: 4.9 miles
room_size: 428.0 square feet
guest_rating: 3.1 stars out of 5
noise_level: 46.0 decibels
age_of_property: 2.0 years<|im_end|>
<|im_start|>assistant
<think>

</think>

{"
\end{verbatim}
}

\noindent The assistant turn is prefixed with \texttt{\{"} to constrain the model's generation to a JSON object. Because the self-report occupies a separate context window, the model has no access to a prior decision or assistant turn from the decision task.

\paragraph{Parsing self-reports.} Generation is stopped at the closing brace of the JSON object (max 100 tokens). A self-report is counted as a parse failure and excluded from analysis if (i)~the output is not valid JSON, (ii)~it does not decode to a dictionary mapping strings to floats, or (iii)~its keys do not exactly match the five expected attribute names. Successfully parsed reports have their values cast to floats. Only successfully parsed reports contribute to the reported preference vector~$\tilde{\bm{p}}$.%

\subsection{Inferring behavioral preferences via logistic regression}
\label{app:behavioral-preferences}

To infer a behavioral preference vector $\hat{\bm{p}}$ from the model's choice behavior, we fit a logistic regression to the model's choice probabilities. For a trial with option~A attributes $\bm{x}^A$ and option~B attributes $\bm{x}^B$, the feature vector is the normalized attribute difference:
\[
    d_i = \frac{\bm{x}^A_i - \bm{x}^B_i}{a_i^{\max} - a_i^{\min}}, \qquad i = 1, \dots, 5.
\]
The model's probability of choosing option~A, denoted $\pi_A$, serves as a soft label. Using soft labels rather than hard-thresholded decisions preserves information about the model's level of uncertainty: a trial where the model assigns 98\% to~A carries more information about its preferences than one where it assigns 52\%. Since standard logistic regression requires hard labels, we use a duplication approach: each trial generates two training rows with the same feature vector $\bm{d}$---one labeled $y = 1$ with sample weight $\pi_A$, and one labeled $y = 0$ with sample weight $1 - \pi_A$. We fit an intercept-free $L_2$-regularized logistic regression ($C = 1.0$, L-BFGS solver) using scikit-learn \citep{pedregosaScikitlearnMachineLearning2011}. The fitted coefficient vector is then rescaled so that $\max_i |\hat{p}_i| = 100$, placing behavioral preferences on the same scale as target preferences.

\subsection{Metric calculations}
\label{app:metric-details}
For each $S, X$, we 
calculate $\smash{\tilde{\bm{p}}^{(S,X)}}$ as the element-wise average over 24 self-report prompts, and $\smash{\hat{\bm{p}}^{(S,X)}}$ as the output of a logistic regression over 16 decisions.

\section{Multi-character adapter training and evaluation details}
\label{app:multi-character-adapter-training-details}

All multi-character adapters are trained using LoRA  with rank-stabilized scaling.
Adapters are applied to all linear projections in each targeted transformer layer: the attention projections ($W_Q$, $W_K$, $W_V$, $W_O$) and the gated MLP projections ($W_\text{gate}$, $W_\text{up}$, $W_\text{down}$).
We use a LoRA rank of 8 with $\alpha = 8$.
Training uses the AdamW optimizer with $\beta_1 = 0.9$, $\beta_2 = 0.99$, $\epsilon = 10^{-8}$, no weight decay, and a maximum gradient norm of $1.0$.
The learning rate is $10^{-4}$ with 1600 warmup steps. Unless otherwise indicated, the scheduler is simply constant with warmup.
All runs train for a single epoch with an effective batch size of 32 and use 8-bit quantization.
Training is performed in BF16 precision with a fixed seed of 2025.

Within a given experiment, all adapters are trained on 100 characters drawn from the same pool.

\subsection{Scaling experiment}
\label{app:scaling-experiment}

To study how model size affects the capacity for introspection, we train multi-character adapters on five Qwen3 model sizes, applying LoRA to all transformer layers.

\begin{table}[h]
\centering
\caption{Qwen3 model configurations used in the scaling experiment.}
\label{tab:scaling-models}
\begin{tabular}{lc}
\toprule
Model & Layers \\
\midrule
Qwen3-0.6B & 28 \\
Qwen3-4B   & 36 \\
Qwen3-8B   & 36 \\
Qwen3-14B  & 40 \\
Qwen3-32B  & 64 \\
\bottomrule
\end{tabular}
\end{table}

\subsection{Tuning 32B}
Our main experiments use a specific adapter trained on top of Qwen3-32B. This was the result of a shallow hyperparameter tune. Starting with the hyperparameters described above, we evaluated 15 random draws from the space of:
\begin{description}
    \item[learning rate] [3e-4, 1e-4, 3e-5]
    \item[weight decay] [0.0, 0.01]
    \item[scheduler] [constant with warmup, cosine with warmup]
    \item[warmup steps] [800, 1600, 3200]
    \item[LoRA rank] [4, 8, 16]
\end{description}

We evaluated using a simple average of decision performance and faithfulness after 2000 training steps. The highest performer mostly kept the starting hyperparameters, but used a cosine scheduler and weight decay of 0.01.

\subsection{Reasoning-mode evaluation}
\label{app:reasoning-mode}

We evaluate the Qwen3-32B training trajectory from Section~\ref{sec:model-organisms} with reasoning enabled at inference time; training itself remains entirely in non-reasoning mode. By the end of training, decision performance and faithfulness are both high and close to their non-reasoning endpoints, and target--report correlation is also high. Across the trajectory, reasoning-mode faithfulness remains comparatively high and does not reproduce the pronounced early collapse observed without reasoning. This is a single-run evaluation without uncertainty bands.

\begin{figure}[t]
    \centering
    \includegraphics[width=\textwidth]{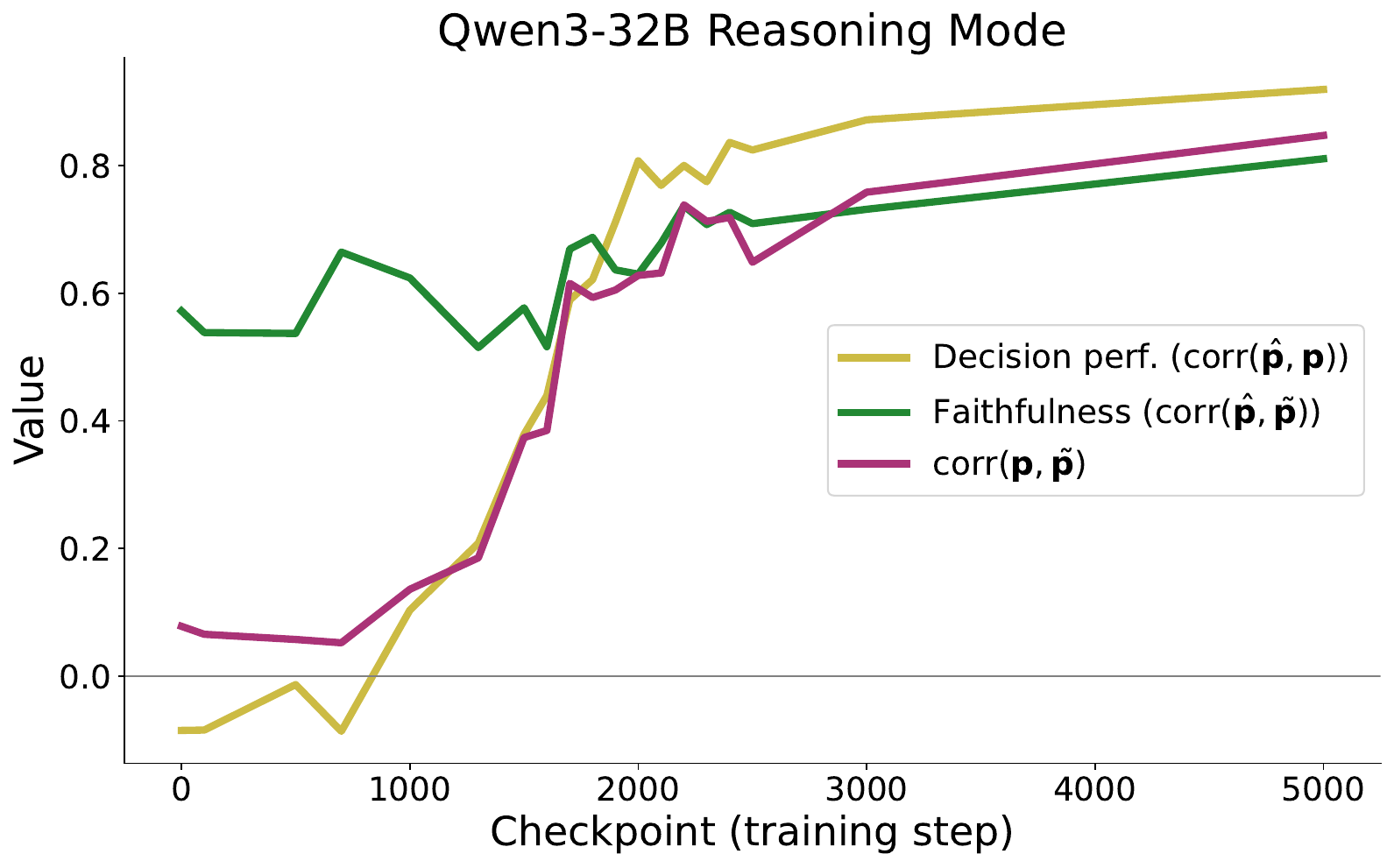}
    \caption{
        \textbf{Qwen3-32B evaluated in reasoning mode.} We evaluate checkpoints trained entirely in non-reasoning mode. Decision performance is $\operatorname{corr}(\hat{\bm{p}},\bm{p})$, faithfulness is $\operatorname{corr}(\hat{\bm{p}},\tilde{\bm{p}})$, and the third curve is $\operatorname{corr}(\bm{p},\tilde{\bm{p}})$. Lines connect the 19 evaluated checkpoints from one run; no uncertainty bands are shown.
    }
    \label{fig:reasoning-mode-trajectory}
\end{figure}

\subsection{Layer freezing experiment}
\label{app:layer-freezing-experiment}

To investigate which layers are important for encoding character preferences, we vary the number of layers that receive LoRA adapters in Qwen3-14B (40 layers total).
In all conditions, LoRA is applied starting from layer 0 up to (but not including) layer $k$, for $k \in \{5, 10, 15, 20, 25, 30, 35\}$.
Layers $k$ and above remain frozen at their pretrained weights.

\subsection{Evaluation details}
We compute decision performance and faithfulness metrics as described in Appendix~\ref{app:metric-details}. In the scaling and layer freezing experiments, we train on 100 characters but, in the interest of efficiency, evaluate on a fixed sample of 32 characters. Figure~\ref{fig:overview}b shows these metrics evaluated on the full set of 100 characters, however.

\subsection{Gemma-4 model-family and scale replication}
\label{app:gemma-replication}

We repeat the decision-task sweep on Gemma-4 E4B and 31B (\texttt{google/gemma-4-E4B-it} and \texttt{google/gemma-4-31B-it}). Among 15 sampled configurations, only 31B runs pair decision performance $\geq0.9$ with faithfulness $\geq0.8$; the remaining runs with strong decision performance have faithfulness $\leq0.3$. Additionally, Gemma-4 does not reproduce Qwen3-32B's within-run delay: the metrics either rise together or faithfulness stays low. We therefore compare separately trained high- and low-faithfulness 31B adapters.

\begin{figure}[p]
    \centering
    \begin{subfigure}[t]{0.48\textwidth}
        \centering
        \includegraphics[width=\textwidth]{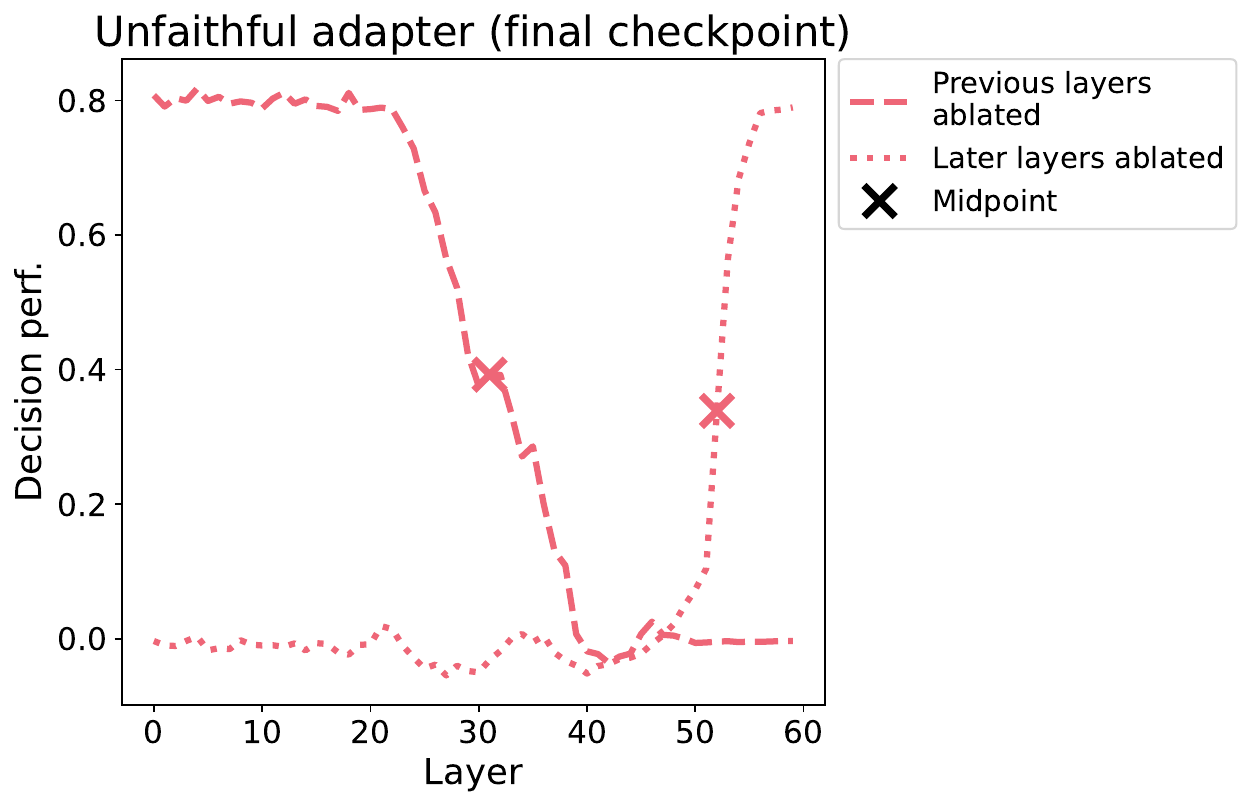}
        \caption{Run selected for low final faithfulness.}
    \end{subfigure}
    \hfill
    \begin{subfigure}[t]{0.48\textwidth}
        \centering
        \includegraphics[width=\textwidth]{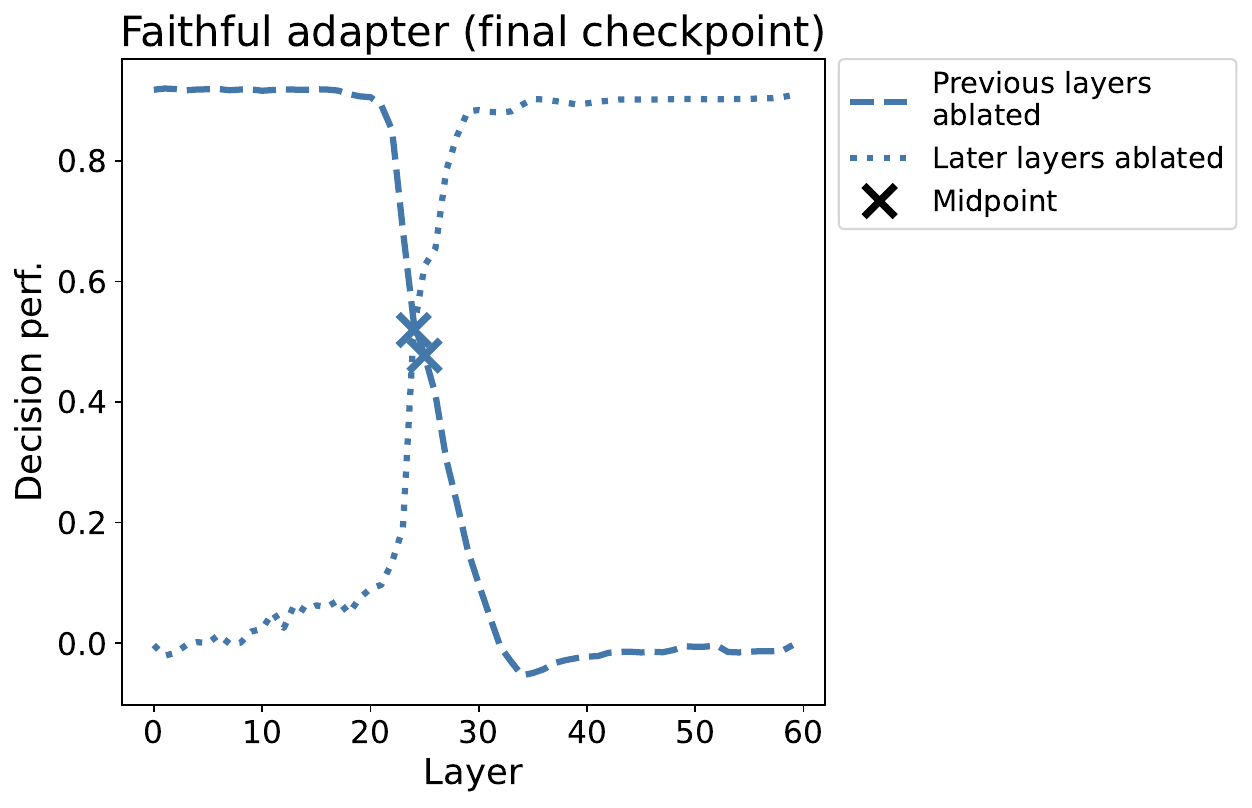}
        \caption{Run selected for high final faithfulness.}
    \end{subfigure}

    \medskip
    \begin{subfigure}[t]{0.62\textwidth}
        \centering
        \includegraphics[width=\textwidth]{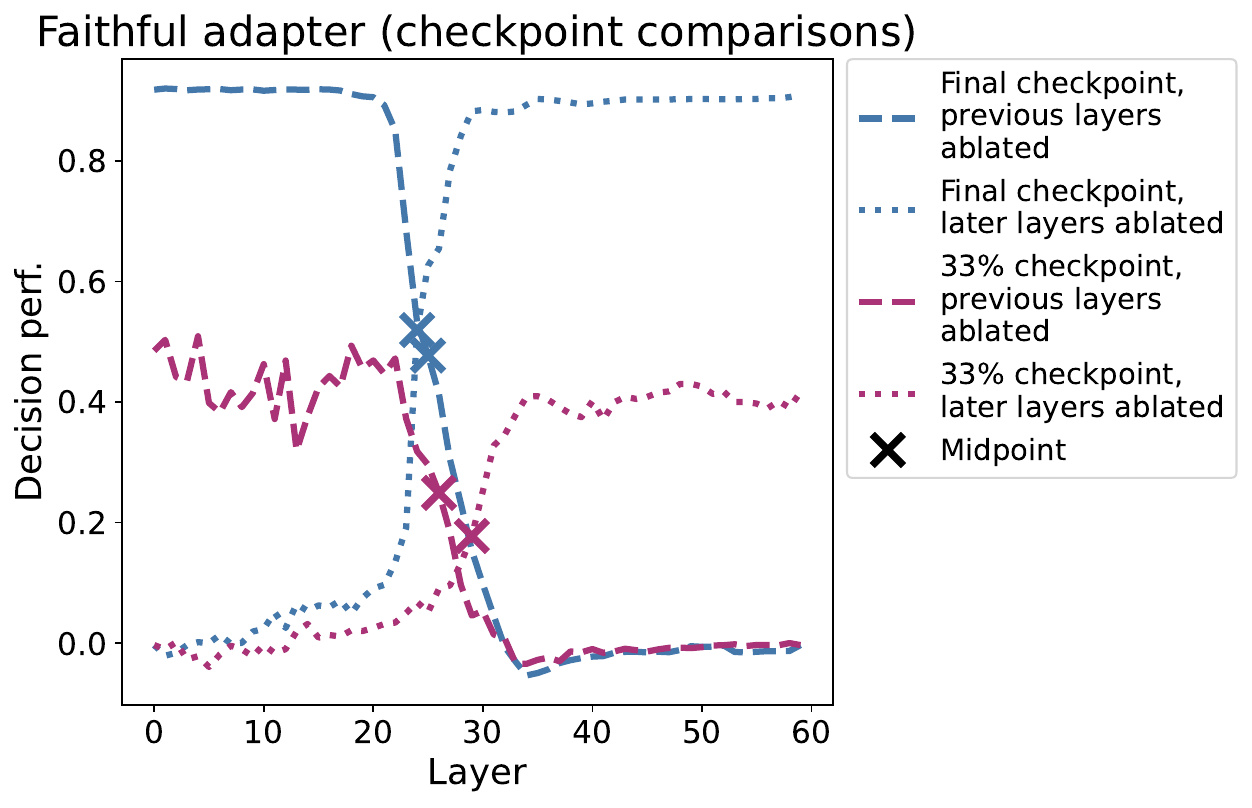}
        \caption{Run selected for high final faithfulness, at one-third and final checkpoints.}
    \end{subfigure}
    \caption{
        \textbf{Gemma-4-31B decision performance under layer ablation.} At boundary $k$, dashed curves ablate layers below $k$ and dotted curves ablate layers above $k$; crosses mark the layer nearest the midpoint between each curve's endpoint values. Panels (a) and (b) show separately trained runs selected for low and high final faithfulness; panel (c) compares two checkpoints from the latter run.
    }
    \label{fig:gemma-layer-ablations}
\end{figure}

Figure~\ref{fig:gemma-layer-ablations} applies the layer-ablation analysis of Section~\ref{sec:weight-ablations-and-frozen-layers}.

\clearpage

\section{Single-character adapter training}
\label{app:lightweight-adapter-training}

Single-character adapters follow the same setup as the multi-character adapters described in Appendix~\ref{app:multi-character-adapter-training-details}, with the following exceptions. We use a LoRA rank of 2 (with $\alpha = 2$), a learning rate of $10^{-3}$, and a cosine scheduler with 20 warmup steps. Each adapter is trained for 24 total steps on data from a single character.

\subsection{Selection-filter robustness}
\label{app:unfiltered-adapters}

We repeat the full attribution-similarity pipeline after removing all three adapter-selection filters from Section~\ref{sec:adapter-training}. We randomly select 10 adapters trained on the early, unfaithful backbone and pair them with the corresponding 10 adapters trained on the later, faithful backbone. The attribution-similarity gap remains positive and significant ($\Delta=0.21$; paired-bootstrap 95\% CI $[0.09, 0.33]$), comparable in magnitude to the filtered estimate of $0.26$, with a wider interval from the smaller sample.

\section{Negative faithfulness}
\label{app:negative-faithfulness}
In Section \ref{sec:model-organisms}, we observed that 4B and 8B models end up with noticeably \emph{negative} faithfulness scores, despite having strong decision performance. How could this be? Figure \ref{fig:14b-unfrozen-layers-comparison} might shed some (partial and \emph{speculative}) light on this question. Our layer freezing experiments show that, when the 14B model is allowed to modify its later layers, it displays a similar pattern of negative faithfulness to the full smaller models. Perhaps in the full 14B model, both faithful early layer and unfaithful later layer representations are present (to various degrees) during training; under this account, 4B and 8B models only contain this second kind of unfaithful representation.

\section{Reversed layer freezing experiment}
\label{app:reverse-freezing-experiment}

The layer freezing experiment of Section~\ref{sec:weight-ablations-and-frozen-layers} (Figure~\ref{fig:14b-unfrozen-layers-comparison}) shows that adapters restricted to the first $k \in \{10, 15, 20\}$ layers of Qwen3-14B have more faithful self-reports than adapters trained on all 40 layers. One concern is that this could simply reflect the smaller total parameter count of the trainable adapter, rather than the location of the trainable layers per se.

\begin{figure}[h]
    \centering
    \includegraphics[width=\textwidth]{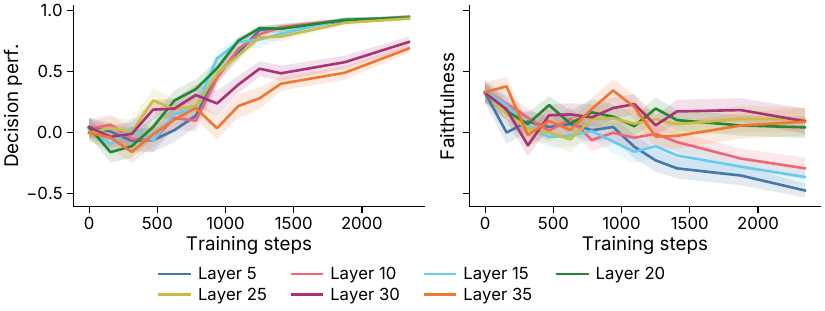}
    \caption{
        \textbf{Qwen3-14B behavior and report when only \emph{late} layers are trainable.} We train adapters on Qwen3-14B, with all layers below $k \in \{5, 10, 15, 20, 25, 30, 35\}$ frozen---the reverse of the configuration in Figure~\ref{fig:14b-unfrozen-layers-comparison}. 
    }
    \label{fig:14b-sweep-min}
\end{figure}

To rule this out, we run a second experiment in which we freeze the \emph{first} $k$ layers, sweeping $k \in \{5, 10, 15, 20, 25, 30, 35\}$. Results are shown in Figure~\ref{fig:14b-sweep-min}. Faithfulness is approximately zero for $k \in \{20, 25, 30, 35\}$ and negative for $k \in \{5, 10, 15\}$---in no case does it match the high levels of faithfulness seen when only early layers are trainable.

\section{Attribution patching}
\label{app:ap-details}
Here, we provide a more detailed description of the attribution patching process used in Section~\ref{sec:ap-setup}.

\paragraph{Attribution scores.}
Each LoRA site $s$ injects an additive perturbation $\bm{\delta}_s = \alpha \cdot B_s A_s \bm{x}_s$ into the output of its host linear layer, where $A_s \in \mathbb{R}^{r \times d_{\mathrm{in}}}$ and $B_s \in \mathbb{R}^{d_{\mathrm{out}} \times r}$ are the low-rank factors, $\alpha = \texttt{lora\_alpha} / r$ is the fixed LoRA scaling coefficient, and $\bm{x}_s$ is the input activation. We compute a per-site, per-output-dimension attribution via the path integral approximation
\begin{equation}
    \bm{a}^s = \sum_t \frac{1}{K} \sum_{k=1}^{K} \bm{\delta}_{s,t}^{(k)} \odot \nabla_{\hat{\bm{\delta}}_{s,t}} \mathcal{L}(\beta_k),
    \label{eq:path-integral}
\end{equation}
where $\mathcal{L}$ is the loss function defined below, $\bm{\delta}_{s,t}^{(k)} = \alpha \cdot B_s A_s \bm{x}_{s,t}^{(k)}$ is the pre-$\beta$ adapter output at site $s$, token position $t$, and integration step $k$, $\hat{\bm{\delta}}_{s,t}^{(k)} = \beta_k \cdot \bm{\delta}_{s,t}^{(k)}$ is the scaled contribution actually injected into the network, and $\beta_k = (k - \tfrac{1}{2})/K$ is the midpoint quadrature point. We use $K = 7$ integration steps and average over 50 inputs for each task type; the estimate stabilizes by four steps (Appendix~\ref{app:ap-step-robustness}). The resulting $\bm{a}^s \in \mathbb{R}^{d_{\mathrm{out}}}$ gives the attribution for site $s$, summed across all token positions.

\subsection{Integration-step robustness}
\label{app:ap-step-robustness}

The main analysis uses $K=7$ integrated-gradient steps. We rerun attribution patching with $K\in\{1,\ldots,6\}$. Table~\ref{tab:ap-step-robustness} shows that the attribution-similarity gap reaches a magnitude comparable to the main estimate by $K=4$; every bootstrap interval for $K\geq4$ excludes zero.

\begin{table}[ht]
    \centering
    \caption{Attribution-similarity difference as the number of integrated-gradient steps varies.}
    \label{tab:ap-step-robustness}
    \begin{tabular}{ccc}
        \toprule
        Steps ($K$) & Mean difference & 95\% bootstrap CI \\
        \midrule
        1 & 0.06 & $[-0.01, 0.13]$ \\
        2 & 0.10 & $[0.00, 0.19]$ \\
        3 & 0.10 & $[0.01, 0.18]$ \\
        4 & 0.31 & $[0.22, 0.41]$ \\
        5 & 0.29 & $[0.20, 0.39]$ \\
        6 & 0.27 & $[0.18, 0.37]$ \\
        7 (main) & 0.26 & $[0.16, 0.36]$ \\
        \bottomrule
    \end{tabular}
\end{table}

\subsection{Loss function}
\label{app:ap-loss-function}
Let $f_\theta$ denote the next-token distribution produced by the model with the adapter at full scale ($\beta = 1$), and $f_{\theta,\beta}$ the distribution with all adapter contributions scaled by $\beta$. We define the loss as the KL divergence between the full-adapter model (teacher) and the scaled model (student), averaged over a set of output-relevant token positions $\mathcal{T}$:
\begin{equation}
    \mathcal{L}(\beta) = \frac{1}{|\mathcal{T}|}\sum_{t \in \mathcal{T}} D_{\mathrm{KL}}\!\bigl(f_\theta(x_{\leq t}) \;\|\; f_{\theta,\beta}(x_{\leq t})\bigr),
\end{equation}
where $f_\theta(x_{\leq t})$ and $f_{\theta,\beta}(x_{\leq t})$ are distributions over the vocabulary $\mathcal{V}$ conditioned on the prefix $x_{\leq t}$.
For decision prompts, $\mathcal{T}$ contains the single position whose next token is the A/B selection; for self-report prompts, $\mathcal{T}$ contains the positions whose next tokens are the numeric values in the generated JSON (see Appendix~\ref{app:kl-positions} for details).

Because the attributions approximate each site's contribution to $\mathcal{L}(1) - \mathcal{L}(0) \leq 0$, the most important sites are those with the most negative scores. (Figure~\ref{fig:attribution-results}c therefore plots the negated score, which we call \emph{importance}.)

\subsection{KL loss positions}
\label{app:kl-positions}

The KL loss in Appendix~\ref{app:ap-loss-function} is evaluated at a task-dependent set of token positions~$\mathcal{T}$.

\paragraph{Decision prompts.} For decision prompts, $\mathcal{T}$ contains a single position: the last token of the prompt, whose next-token prediction determines the A/B selection.

\paragraph{Self-report prompts.} For self-report prompts, the model generates a JSON object mapping attribute names to numeric weights, e.g.:

{\small
\begin{verbatim}
{"distance_to_city_center": -50, "room_size": 100, ...}
\end{verbatim}
}

\noindent The set $\mathcal{T}$ contains the positions that predict each token in the numeric values. Qwen3's tokenizer encodes each digit as a separate token, so a value such as \texttt{-50} is tokenized as three tokens: \texttt{-}, \texttt{5}, \texttt{0}. Concretely, a numeric span includes:
\begin{itemize}
    \item an optional leading space or \texttt{ -} (space-hyphen) token, which distinguishes positive from negative values;
    \item one token per digit;
    \item an optional decimal point and subsequent digit tokens.
\end{itemize}
For each such span, $\mathcal{T}$ contains the positions that \emph{predict} these tokens (i.e., shifted back by one), as well as the position immediately after the final digit. Including this trailing position is important because the logits there distinguish, e.g., whether the model predicts that the value \texttt{10} continues as \texttt{100} or terminates with a comma.

\section{Attribution patching across characters in a single backbone}
\label{app:backbone-ap}

Section~\ref{sec:ap-results} compares two \emph{populations} of single-character adapters each trained on top of a different multi-character backbone. Here we run a complementary analysis in which faithfulness varies across characters, \emph{within} a single backbone without any lightweight adapter on top. Whereas the main experiment asks whether attribution similarity separates faithful and unfaithful \emph{models}, this experiment asks whether, fixing a single model, attribution similarity tracks the characters it happens to self-report most faithfully on.

\paragraph{Setup.}
We use checkpoint~1000 of the 32B multi-character adapter from Section~\ref{sec:model-organisms}---the same adapter that serves as the early, unfaithful backbone in the main experiment. Although its mean faithfulness is low (around $0.25$), per-character faithfulness varies considerably, giving us the spread needed for a within-backbone correlation.

For each of 100 characters, we compute decision-task and self-report-task attribution vectors directly on the multi-character adapter using the procedure of Appendix~\ref{app:ap-details}, and take the cosine similarity between them as that character's attribution similarity. Per-character faithfulness is computed as in the main paper.

\begin{figure}[h]
    \centering
    \includegraphics[width=0.6\textwidth]{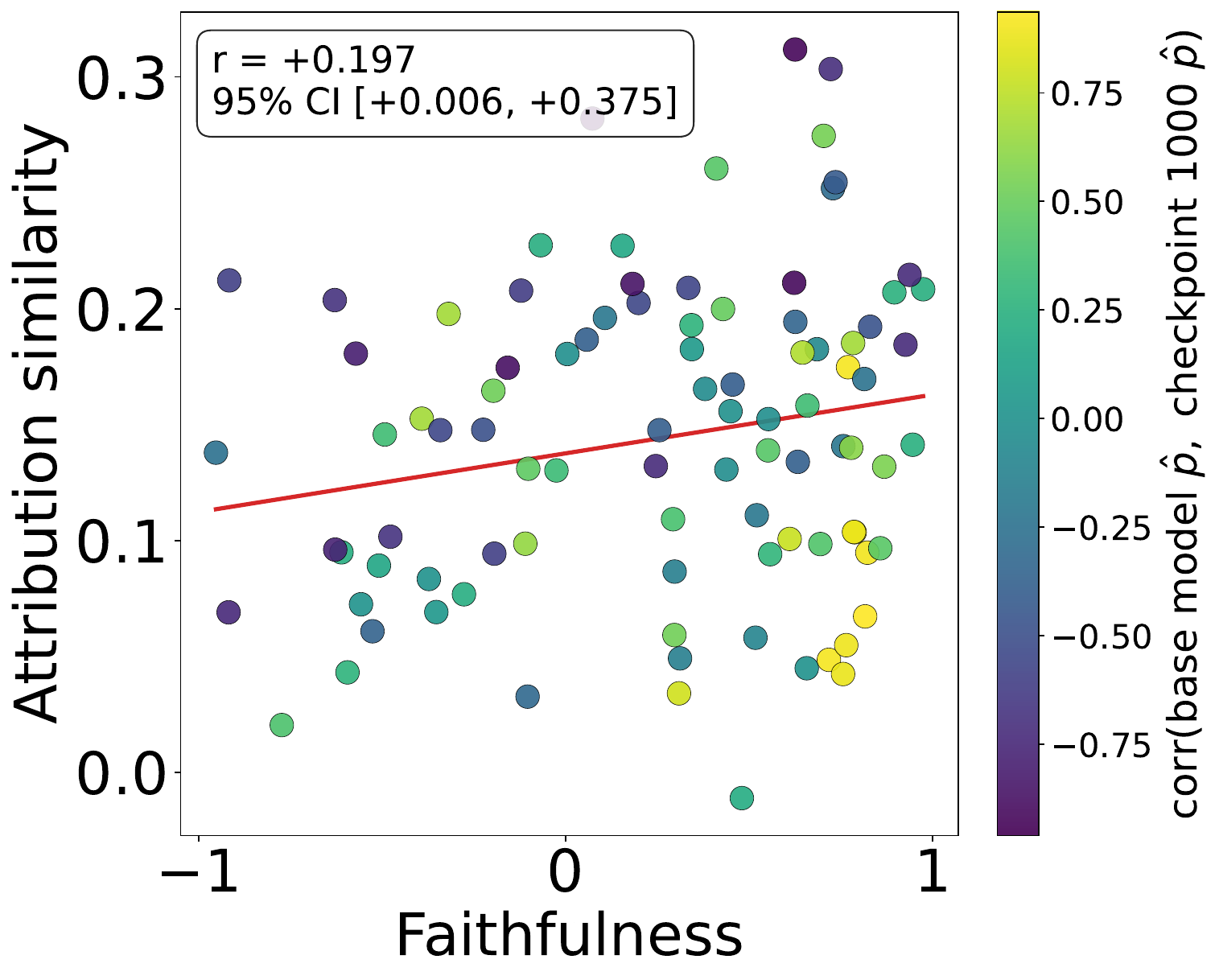}
    \caption{
        \textbf{Per-character attribution similarity vs. faithfulness within a single backbone.} Each point is one of 100 characters, evaluated against checkpoint~1000 of the 32B multi-character adapter (the early, unfaithful backbone; no lightweight adapter on top). Across characters, attribution similarity correlates weakly but positively with faithfulness (Pearson $r = 0.197$, bootstrap 95\% CI $[0.006, 0.375]$). Color encodes the correlation, for that character, between the behavioral preferences inferred from the base 32B model and those inferred from checkpoint~1000.
    }
    \label{fig:backbone-ap-scatter}
\end{figure}

\paragraph{Result.}
Figure~\ref{fig:backbone-ap-scatter} plots attribution similarity against faithfulness across the 100 characters. We find a weak but nonzero positive relationship (Pearson $r = 0.197$; bootstrap 95\% CI $[0.006, 0.375]$). The direction matches the main experiment's prediction---characters that the model self-reports more faithfully on are also those whose decision and self-report computations share more weight---but the effect is small, and the lower edge of the confidence interval barely excludes zero. We see this as a sign of life rather than a standalone finding, and we leave a more thorough investigation to future work.

\section{AI usage statement}
The authors used AI to assist with implementing and editing the paper.

\end{document}